\documentclass[10pt,twocolumn,letterpaper]{article}

\usepackage{cvpr}              
\definecolor{cvprblue}{rgb}{0.21,0.49,0.74}
\usepackage[pagebackref,breaklinks,colorlinks,allcolors=cvprblue]{hyperref}
\usepackage{multirow}
\def\paperID{6756} 
\def\confName{CVPR}
\def\confYear{2026}

\title{MODIX: A Training-Free Multimodal Information-Driven Positional Index Scaling for Vision-Language Models}

\author{
Ruoxiang Huang$^{*\dagger}$ \quad Zhen Yuan$^{*\ddagger}$\\
Peking University, China\\
{\tt\small \{15711215564hrx, zhenyuanyz1219\}@gmail.com}\\
}
\begin{document}
\maketitle
{
\renewcommand{\thefootnote}{}  
\footnotetext{*Equal contribution.}
\footnotetext{$^{\dagger}$Incoming graduate student.}
\footnotetext{$^{\ddagger}$Work done during an internship.}
}
\begin{abstract}
Vision-Language Models (VLMs) have achieved remarkable progress in multimodal understanding, yet their positional encoding mechanisms remain suboptimal. Existing approaches uniformly assign positional indices to all tokens, overlooking variations in information density within and across modalities, which leads to inefficient attention allocation where redundant visual regions dominate while informative content is underrepresented. We identify positional granularity as an implicit resource and propose MODIX (Multimodal Information-Driven Positional IndeX Scaling), a training-free framework that dynamically adapts positional strides based on modality-specific contributions. MODIX jointly models intra-modal density via covariance-based entropy and inter-modal interaction via cross-modal alignment to derive unified scores, which rescale positional indices to allocate finer granularity to informative modalities while compressing redundant ones, without requiring any modification to model parameters or architecture. Experiments across diverse architectures and benchmarks demonstrate that MODIX consistently improves multimodal reasoning and adaptively reallocates attention according to task-dependent information distributions, suggesting that positional encoding should be treated as an adaptive resource in Transformers for multimodal sequence modeling.
\end{abstract}    
\section{Introduction}

Vision-Language Models (VLMs) have revolutionized multimodal understanding by integrating visual perception with natural language processing~\cite{radford2021clip,alayrac2022flamingo,liu2023llava}. At the heart of these models lies the Transformer architecture~\cite{vaswani2017attention}, whose self-attention mechanism fundamentally relies on positional encoding to capture sequential relationships and spatial structure across modalities. However, current positional encoding mechanisms in VLMs remain fundamentally uniform, treating all tokens identically regardless of their information content or cross-modal importance.

This uniform treatment stands in stark contrast to the heterogeneous nature of multimodal data. Text tokens are semantically dense, with each word contributing distinct information, while visual tokens derived from fixed-size image patches often exhibit substantial spatial redundancy in uniform backgrounds or repetitive textures. This information density asymmetry manifests both between and within modalities, yet existing schemes apply identical sequential indices $p_i = i$ to all tokens, wasting representational capacity on redundant content while under-representing information-rich regions. Moreover, modality contributions vary dramatically across tasks, from visual-dominant scene understanding to text-dominant chart interpretation. Current VLMs employ static positional encoding that fails to account for these task-dependent shifts and inter-modal dependencies.

Rotary Position Embedding (RoPE)~\cite{su2024roformer}, the predominant mechanism in modern VLMs~\cite{liu2023llava,wang2025internvl,bai2025qwen3vltechnicalreport}, encodes relative positions through rotations based on token distance $\Delta p = p_j - p_i$. While effective for homogeneous text, RoPE's uniform stride assignment treats semantically rich text tokens and redundant background patches with identical positional granularity, providing uniform discriminative power regardless of information content.

We introduce MODIX (Multimodal Information-Driven Positional Index Scaling), a training-free framework that dynamically adapts positional strides based on information-theoretic analysis. Our key insight is that positional granularity should reflect information contribution: modalities with higher information density deserve finer positional distinctions, while redundant content tolerates coarser spacing. Operating purely at inference time, MODIX requires no retraining, parameter updates, or architectural modifications, enabling plug-and-play integration with existing RoPE-based VLMs.

MODIX quantifies modality contributions through two complementary components: first, it estimates intra-modal information density via covariance-based entropy; second, it measures inter-modal interaction strength via cross-modal alignment. This dual analysis captures the synergistic nature of multimodal understanding, where a modality's contribution depends on both its internal information richness and its interaction with other modalities. Grounded in the principle that denser information warrants finer positional resolution, MODIX computes adaptive strides following the principle $\Delta_m \propto 1/\tilde{C}_m$ and reconstructs adjusted positional indices $\mathbf{P}'$ that seamlessly integrate into standard RoPE.

We evaluate MODIX across three VLM architectures (Qwen3-VL~\cite{bai2025qwen3vltechnicalreport}, InternVL3.5~\cite{wang2025internvl}, LFM2-VL~\cite{amini2025lfm2technicalreport}) spanning 1B to 8B parameters and seven diverse benchmarks (ScienceQA~\cite{lu2022scienceqa}, RealWorldQA~\cite{xai2024realworldqa}, DocVQA~\cite{mathew2021docvqa}, ChartQA~\cite{masry2022chartqa}, AI2D~\cite{kembhavi2016ai2d}, BLINK~\cite{fu2024blink} and Video-MME~\cite{fu2025video}), demonstrating consistent improvements. Our analysis reveals that MODIX automatically adapts to task characteristics, dynamically adjusting positional granularity to match information distribution.

Our contributions are: (1) We formalize the information asymmetry problem in multimodal positional encoding. (2) We introduce an information-theoretic framework modeling both intra-modal density and inter-modal interaction. (3) We propose MODIX, which instantiates this framework as a training-free, inference-time method for dynamic positional index scaling in VLMs. (4) We provide comprehensive validation across diverse architectures and benchmarks.
\section{Related Work}

\subsection{Vision-Language Models}

Vision-Language Models bridge visual and textual modalities through joint representation learning. Following Vision Transformers~\cite{dosovitskiy2021vit}, modern VLMs~\cite{alayrac2022flamingo,li2023blip2,liu2023llava,wang2025internvl,bai2025qwen3vltechnicalreport} encode images as sequences of patches, concatenating them with text tokens in a unified embedding space through sequential positional encoding. However, this uniform treatment overlooks fundamental differences in information density between modalities. MODIX addresses this limitation through training-free, information-driven positional index adaptation.

\subsection{Positional Encoding in Transformers}

Positional encoding has evolved from absolute~\cite{vaswani2017attention,devlin2019bert} to relative paradigms~\cite{dai2019transformerxl,raffel2020t5}. Rotary Position Embedding (RoPE)~\cite{su2024roformer} encodes relative positions through geometric rotations, becoming standard in modern VLMs~\cite{liu2023llava,wang2025internvl}. Recent adaptive mechanisms~\cite{press2022alibi,chen2023extendingcontextwindowlarge,peng2024yarn} adjust attention biases or frequency parameters for improved extrapolation. However, \textit{all existing methods apply uniform positional strides} ($p_i = i$) \textit{regardless of token information content}. To our knowledge, MODIX is the first training-free method to dynamically adapt positional granularity based on information-theoretic analysis of modality contributions.

Recent multimodal positional encoding variants have explored different design objectives. V2PE~\cite{Ge2025V2PE} improves multimodal long-context capability through variable visual position encoding, CircleRoPE~\cite{wang2025circle} mitigates cross-modal positional bias, and MHRoPE~\cite{huang2025revisiting} revisits multimodal RoPE design from an architectural perspective, proposing training-time principles such as positional coherence and frequency utilization to improve modality alignment. Unlike these approaches, MODIX requires no retraining and adapts positional granularity dynamically based on task-dependent information contribution rather than fixed design rules.

\subsection{Information Theory in Multimodal Learning}

Information-theoretic frameworks guide multimodal representation learning through principles such as the Information Bottleneck~\cite{tishby2000information,federici2020multiview} and Mutual Information maximization~\cite{radford2021clip,bachman2019amdim,oord2018cpc}. Recent work applies information theory to modality fusion~\cite{hazarika2020misa,han2023trusted} and heterogeneity analysis~\cite{liang2022quantifying}. However, systematic application of these principles to the design of positional strides in multimodal models remains unexplored. Token pruning methods~\cite{bolya2023tome,rao2021dynamicvit,yin2022avit} physically remove redundant tokens but require architectural modifications and lose spatial information. MODIX instead adjusts positional spacing through principled information-theoretic analysis while preserving all tokens, enabling training-free deployment on pretrained VLMs.
\section{Methodology}
\subsection{Problem Formulation}

Consider a multimodal input with $n_t$ text tokens 
$X_{\text{text}} = \{x_1^t, x_2^t, \ldots, x_{n_t}^t\}$ and $n_v$ vision tokens 
$X_{\text{vision}} = \{x_1^v, x_2^v, \ldots, x_{n_v}^v\}$.
Let $\mathcal{M} = \{\text{text}, \text{vision}\}$ denote the modalities. A VLM projects both into a unified embedding space:
\begin{equation}
\mathbf{E} = [\mathbf{E}_{\text{text}};\; \mathbf{E}_{\text{vision}}] \in \mathbb{R}^{N \times d}
\end{equation}
where $\mathbf{E}_{\text{text}} \in \mathbb{R}^{n_t \times d}$, 
$\mathbf{E}_{\text{vision}} \in \mathbb{R}^{n_v \times d}$, 
$N = n_t + n_v$, $d$ is the embedding dimension, and $[\,\cdot\,;\,\cdot\,]$ denotes concatenation.

Standard VLMs assign uniform positional indices 
$\mathbf{P} = [p_0, p_1, \ldots, p_{N-1}]$ with $p_i = i$, processed by RoPE~\cite{su2024roformer}:
\begin{equation}
\label{eq:standard_rope}
\mathbf{q}_i = \mathbf{R}(\boldsymbol{\theta}, p_i)\,\mathbf{q}_i^{\text{base}}, \quad
\mathbf{k}_j = \mathbf{R}(\boldsymbol{\theta}, p_j)\,\mathbf{k}_j^{\text{base}}
\end{equation}
where $\mathbf{R}(\boldsymbol{\theta}, p) \in \mathbb{R}^{d \times d}$ is the rotation matrix and $\mathbf{q}_i^{\text{base}}, \mathbf{k}_j^{\text{base}} \in \mathbb{R}^d$ are the query and key vectors.

This uniform treatment overlooks modality asymmetry: text tokens convey semantically distinct information with positional encodings optimized during LM pretraining, whereas vision tokens often exhibit substantial spatial redundancy. This information density asymmetry renders uniform positional granularity suboptimal.

\begin{figure*}[t]
    \centering
    \includegraphics[width=0.95\textwidth]{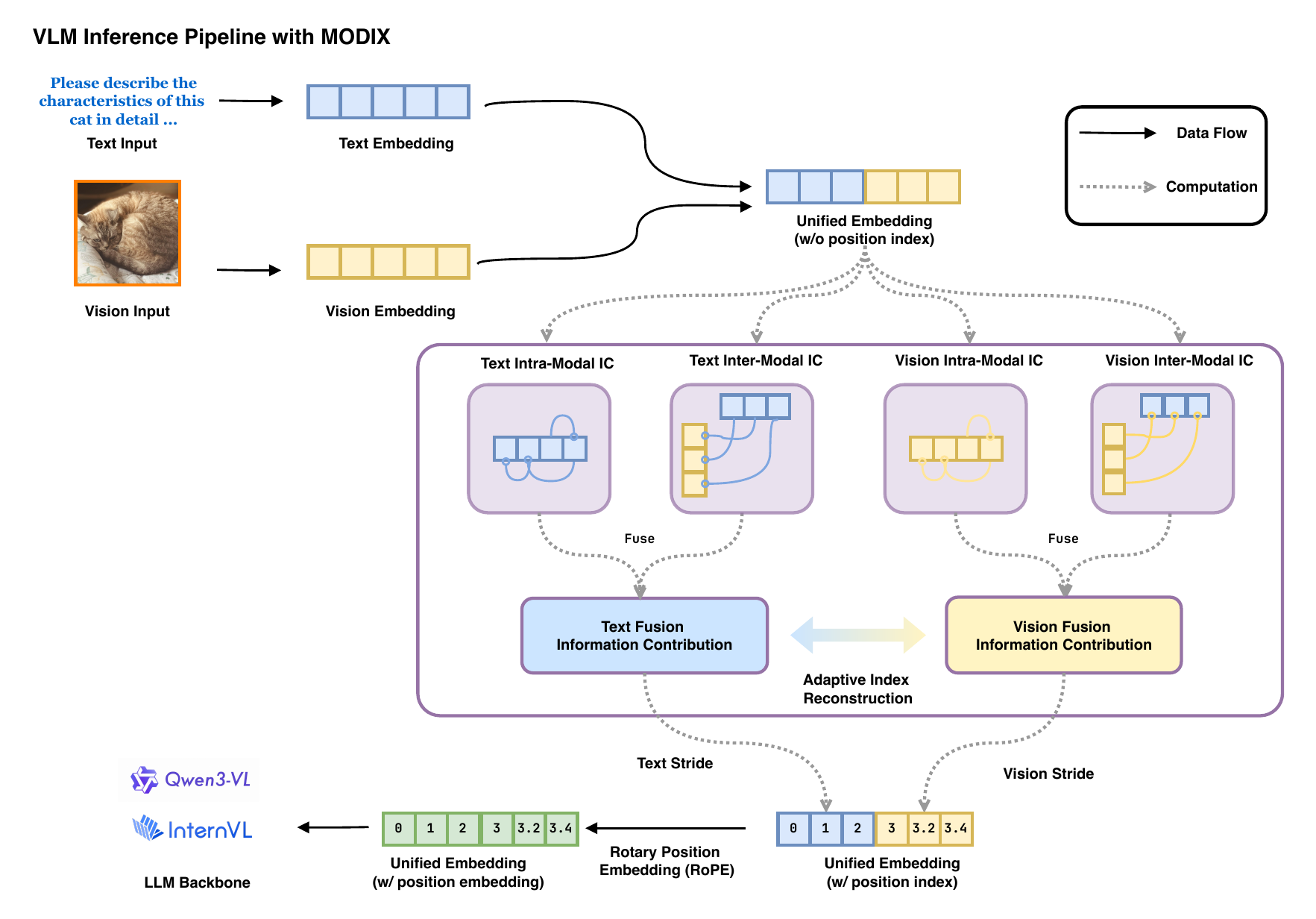}
    \caption{\textbf{MODIX framework.} Dual pathways analyze multimodal embeddings $\mathbf{E}$ to compute information contributions $\tilde{C}_m$, which determine adaptive vision strides while text maintains unit stride. Adjusted indices $\mathbf{P}'$ integrate directly into RoPE without parameter updates.}
    \label{fig:modix_pipeline}
\end{figure*}

We design a training-free approach that adjusts vision strides while preserving text positions. We seek a mapping $f\colon \mathbb{R}^{N \times d} \to \mathbb{R}^N$ producing modality-aware indices:
\begin{equation}
\label{eq:adjusted_indices}
\mathbf{P}' = f(\mathbf{E}) = [p_0',\, p_1',\, \ldots,\, p_{N-1}'] \quad \text{with } p_0' = 0
\end{equation}
such that vision tokens with lower information contribution receive coarser positional granularity, text tokens preserve their original indices, and monotonicity $p_i' < p_j'$ for $i < j$ is maintained. Position encoding then becomes:
\begin{equation}
\label{eq:adjusted_rope}
\mathbf{q}_i' = \mathbf{R}(\boldsymbol{\theta}, p_i')\,\mathbf{q}_i^{\text{base}}, \quad
\mathbf{k}_j' = \mathbf{R}(\boldsymbol{\theta}, p_j')\,\mathbf{k}_j^{\text{base}}
\end{equation}
Attention scores now depend on the adjusted relative distance $|p_i' - p_j'|$, jointly reflecting token separation and modality contribution.

As shown in Figure~\ref{fig:modix_pipeline}, MODIX analyzes $\mathbf{E}$ through dual pathways: an intra-modal pathway quantifies information density within each modality, and an inter-modal pathway captures cross-modal interactions. These pathways produce adaptive vision strides for reconstructing $\mathbf{P}'$, which integrates into RoPE without modifying any model parameters.

\subsection{Information Contribution}
\label{sec:contribution}

To determine how positional granularity should be allocated across modalities, we estimate each modality's contribution from two complementary perspectives: intra-modal information density and inter-modal interaction strength. For each modality $m \in \mathcal{M}$, we compute an intra-modal score $I_m^{\text{intra}}$ and an inter-modal score $I_m^{\text{inter}}$, then fuse them into a unified contribution $\tilde{C}_m$.

\textbf{Intra-Modal Information Contribution.}
We quantify intra-modal information density through covariance-based entropy estimation of the embedding distribution.
For modality $m \in \mathcal{M}$ with embeddings
$E_m = \{\mathbf{e}_1^m, \ldots, \mathbf{e}_{n_m}^m\}$, $\mathbf{e}_i^m \in \mathbb{R}^d$,
we compute the empirical mean
$\bar{\mathbf{e}}_m = \frac{1}{n_m}\sum_{i=1}^{n_m}\mathbf{e}_i^m$
and centered embeddings:
\begin{equation}
\label{eq:centered}
\tilde{\mathbf{e}}_i^m = \mathbf{e}_i^m - \bar{\mathbf{e}}_m, \quad i = 1, \ldots, n_m
\end{equation}
The empirical covariance matrix is then given by:
\begin{equation}
\label{eq:covariance}
\Sigma_m = \frac{1}{n_m}\sum_{i=1}^{n_m}
\tilde{\mathbf{e}}_i^m\,(\tilde{\mathbf{e}}_i^m)^\top \in \mathbb{R}^{d \times d}
\end{equation}
We estimate information density via differential entropy~\cite{cover2006elements}, adopting a Gaussian approximation for computational tractability. This choice is justified by the high dimensionality of the embedding space ($d \approx 1000$) and the regularizing effect of contrastive training objectives~\cite{radford2021clip}. We note that the covariance determinant $\det(\Sigma_m)$ itself provides a distribution-agnostic measure of embedding variability, effectively capturing information volume regardless of the precise distributional form. The entropy proxy is computed as:
\begin{equation}
\label{eq:entropy}
H_m^{\text{intra}} = \frac{1}{2}\log\det(\Sigma_m + \epsilon I_d)
\end{equation}
where $\epsilon = 10^{-6}$ ensures numerical stability. Larger determinant values indicate greater information richness. Normalizing across modalities yields the intra-modal contribution:
\begin{equation}
\label{eq:intra_norm}
I_m^{\text{intra}} =
\frac{H_m^{\text{intra}}}
     {\sum_{m' \in \mathcal{M}} H_{m'}^{\text{intra}}},
\quad \forall\, m \in \mathcal{M}
\end{equation}

\textbf{Inter-Modal Information Contribution.}
A modality's contribution depends not only on its internal richness but also on how strongly it interacts with the other modality. We quantify this interaction in the shared embedding space. We first L2-normalize the embeddings as $\hat{\mathbf{e}}_i^m = \mathbf{e}_i^m / \|\mathbf{e}_i^m\|_2$ and compute the cross-modal similarity matrix:
\begin{equation}
\label{eq:similarity}
S = \hat{E}_{\text{text}}\,\hat{E}_{\text{vision}}^\top
  \in \mathbb{R}^{n_t \times n_v}
\end{equation}
where $S_{ij} = \langle \hat{\mathbf{e}}_i^{\text{text}},\,
\hat{\mathbf{e}}_j^{\text{vision}} \rangle$.
We then define directional interaction scores by averaging the maximum similarity of each token to the opposite modality:
\begin{equation}
\label{eq:t2v}
S_{\text{text}\to\text{vision}}^{\text{inter}}
= \frac{1}{n_t}\sum_{i=1}^{n_t}\max_{j \in [1,n_v]} S_{ij}
\end{equation}
\begin{equation}
\label{eq:v2t}
S_{\text{vision}\to\text{text}}^{\text{inter}}
= \frac{1}{n_v}\sum_{j=1}^{n_v}\max_{i \in [1,n_t]} S_{ij}
\end{equation}
These two quantities capture asymmetric cross-modal dependencies: the first measures how well each text token finds supporting visual evidence, while the second measures how well each visual token aligns with textual semantics.
For unified notation, let
$S_{\text{text}}^{\text{inter}} = S_{\text{text}\to\text{vision}}^{\text{inter}}$
and
$S_{\text{vision}}^{\text{inter}} = S_{\text{vision}\to\text{text}}^{\text{inter}}$.
Normalizing across modalities:
\begin{equation}
\label{eq:inter_norm}
I_m^{\text{inter}} =
\frac{S_m^{\text{inter}}}
     {\sum_{m' \in \mathcal{M}} S_{m'}^{\text{inter}}},
\quad \forall\, m \in \mathcal{M}
\end{equation}

\begin{figure*}[t]
    \centering
    \includegraphics[width=0.95\textwidth]{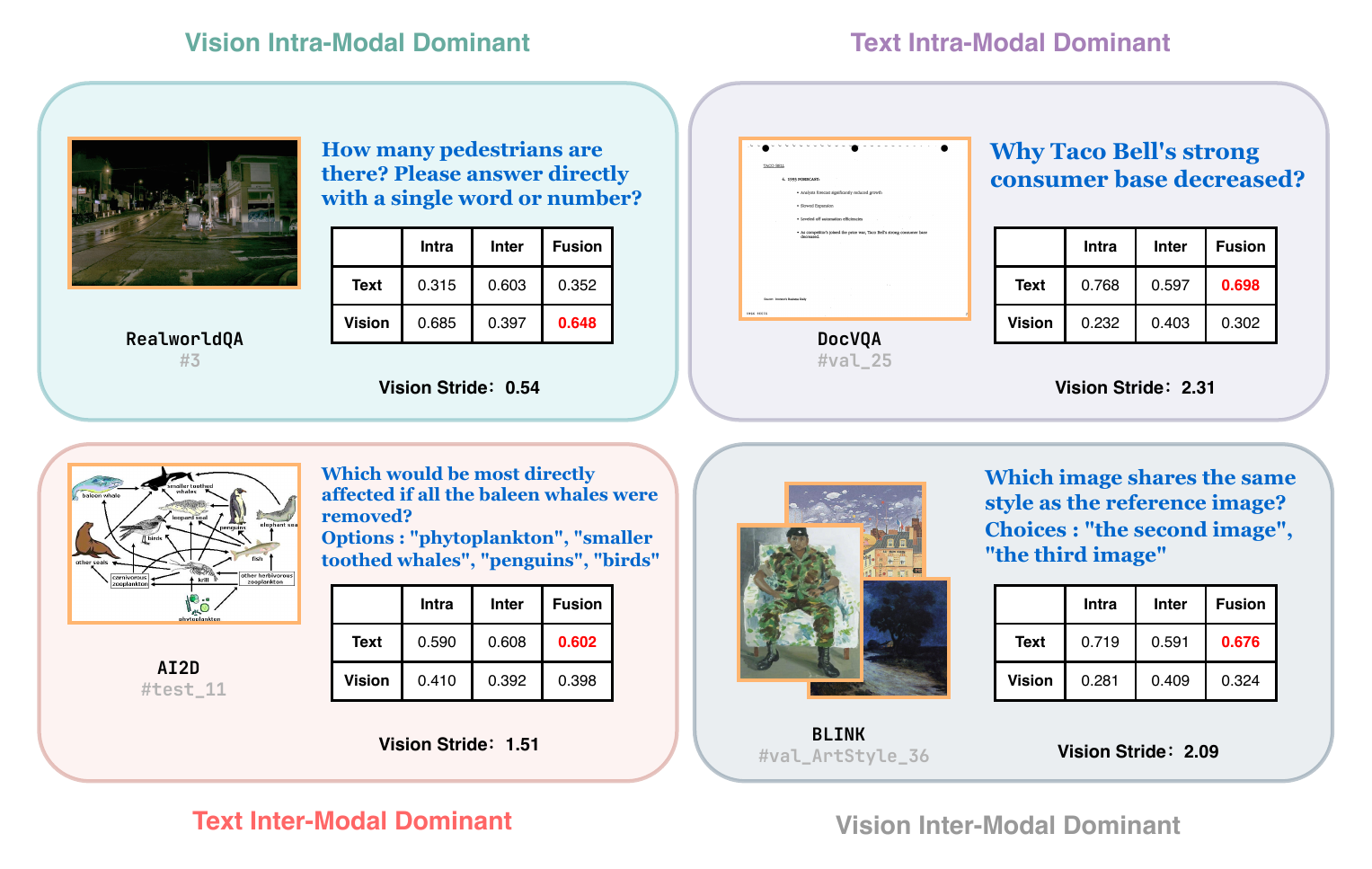}
    \caption{\textbf{Case analysis of MODIX} across four representative task types. Each panel reports the decomposed information contributions (intra-modal, inter-modal, and fused via Eq.~\ref{eq:fusion}) together with the resulting adaptive vision stride. MODIX assigns finer visual granularity when vision contribution dominates and coarser spacing when text dominates.}
    \label{fig:case}
\end{figure*}

\textbf{Contribution Fusion.}
We combine intra-modal and inter-modal contributions using a geometric mean:
\begin{equation}
\label{eq:fusion}
C_m = \left(I_m^{\text{intra}}\right)^{\alpha}
      \left(I_m^{\text{inter}}\right)^{1-\alpha},
\quad \alpha \in [0,1]
\end{equation}
This formulation ensures that a modality achieves high contribution only when it simultaneously exhibits rich internal information and strong cross-modal coherence. Normalizing yields the final unified contribution:
\begin{equation}
\label{eq:final_contribution}
\tilde{C}_m =
\frac{C_m}{\sum_{m' \in \mathcal{M}} C_{m'}},
\quad \forall\, m \in \mathcal{M}
\end{equation}
The resulting $\tilde{C}_{\text{text}}$ and $\tilde{C}_{\text{vision}}$
are used in Section~\ref{sec:scaling} to determine contribution-adaptive positional strides. The choice of $\alpha$ is analyzed in Section~5.1, and Figure~\ref{fig:case} illustrates how these contributions translate into instance-specific stride adaptation across diverse task types.

\subsection{Adaptive Stride Scaling}
\label{sec:scaling}

\textbf{Theoretical Motivation.}
RoPE encodes positional information through rotation matrices applied to query and key vectors, making attention scores fundamentally dependent on relative positional distance~\cite{su2024roformer}. For tokens $i$ and $j$ with positional indices $p_i$ and $p_j$, the attention score takes the form
\begin{equation}
\label{eq:rope_attn}
\text{Attn}(i,j) \propto
(\mathbf{q}_i^{\text{base}})^\top
\mathbf{R}(\boldsymbol{\theta},\, p_j - p_i)\,
\mathbf{k}_j^{\text{base}},
\end{equation}
where $\mathbf{R}(\boldsymbol{\theta}, \Delta p) \in \mathbb{R}^{d \times d}$ is a block-diagonal rotation matrix parameterized by frequency vector $\boldsymbol{\theta}$ and relative offset $\Delta p = p_j - p_i$, with each $2\times2$ block given by
{\small
\begin{equation}
\label{eq:rope_block}
\mathbf{R}_k(\boldsymbol{\theta}, \Delta p) =
\begin{bmatrix}
\cos(k\theta \cdot \Delta p) & -\sin(k\theta \cdot \Delta p)\\
\sin(k\theta \cdot \Delta p) & \cos(k\theta \cdot \Delta p)
\end{bmatrix},
\end{equation}
}%
where $k = 1,\ldots,d/2$ indexes each frequency component.

The rotation angle $k\theta \cdot \Delta p$ grows linearly with positional distance, inducing an attention decay pattern where tokens farther apart receive progressively lower attention weights.

The stride $\Delta_m$ controls the positional index increment between adjacent tokens within modality $m$ and thus determines the effective positional span that modality occupies in RoPE space. Under the softmax normalization constraint $\sum_{j=0}^{N-1}\text{Attn}(i,j)=1$, attention constitutes a fixed budget per query token. Since attention bandwidth is inversely related to positional span under RoPE's distance-dependent decay~\cite{su2024roformer}, a modality with smaller stride receives proportionally higher aggregate attention. To align attention allocation with information contribution, we require the bandwidth ratio between modalities to match their contribution ratio, i.e., $A_{\text{text}}^{\text{total}}/A_{\text{vision}}^{\text{total}} \approx \tilde{C}_{\text{text}}/\tilde{C}_{\text{vision}}$, where $A_m^{\text{total}}$ denotes the aggregate attention received by modality~$m$. Combining with the inverse bandwidth-stride relationship $A_{\text{text}}^{\text{total}}/A_{\text{vision}}^{\text{total}} \propto \Delta_{\text{vision}}/\Delta_{\text{text}}$ and fixing $\Delta_{\text{text}} = 1$ to preserve pretrained textual positional structure, we obtain the adaptive visual stride:
\begin{equation}
\label{eq:stride}
\Delta_{\text{vision}} = \frac{\tilde{C}_{\text{text}}}{\tilde{C}_{\text{vision}}}
\end{equation}
When $\tilde{C}_{\text{vision}}$ is small relative to $\tilde{C}_{\text{text}}$, we have $\Delta_{\text{vision}} > 1$, yielding coarser positional spacing for visual tokens; when $\tilde{C}_{\text{vision}}$ dominates, $\Delta_{\text{vision}} < 1$ produces finer spacing.

We fix $\Delta_{\text{text}}=1$ because text and vision play fundamentally asymmetric roles in VLMs. The language backbone is pretrained with specific positional encoding patterns, under which it learns syntactic, semantic, and discourse-level dependencies intrinsically tied to the sequential index structure~\cite{su2024roformer,press2022alibi}. Altering textual stride would perturb these learned relationships. Visual tokens, by contrast, are introduced through a projection interface from visual representations~\cite{liu2023llava} and lack analogous pretrained positional priors, making their stride a natural degree of freedom for inference-time adaptation.

\textbf{Stride Computation and Index Reconstruction.}
Using the adaptive stride from Equation~\ref{eq:stride}, we reconstruct the adjusted indices $\mathbf{P}' = \{p_i'\}_{i=0}^{N-1}$ in a piecewise manner:
\begin{equation}
\label{eq:reconstruction}
p_i' =
\begin{cases}
i, & i < n_t,\\[2pt]
p_{n_t-1}' + \Delta_{\text{vision}}\cdot(i - n_t + 1), & i \geq n_t,
\end{cases}
\end{equation}
where $N = n_t + n_v$. Text tokens retain their original indices $p_i' = i$ for $i \in [0,\, n_t{-}1]$, while vision tokens begin at $p_{n_t}' = n_t$ and advance with constant stride $\Delta_{\text{vision}}$. Since $\Delta_{\text{vision}} > 0$, the sequence is strictly monotonic. The reconstruction requires only a linear pass over the token sequence. We replace the original indices with $\mathbf{P}'$ when applying RoPE:
\begin{equation}
\label{eq:adjusted_rope_final}
\mathbf{q}_i' = \mathbf{R}(\boldsymbol{\theta}, p_i')\,\mathbf{q}_i^{\text{base}}, \quad
\mathbf{k}_j' = \mathbf{R}(\boldsymbol{\theta}, p_j')\,\mathbf{k}_j^{\text{base}}
\end{equation}
Accordingly, RoPE operates on the adjusted relative offsets $p_j' - p_i'$, yielding a contribution-adaptive positional geometry for multimodal attention. MODIX introduces no parameter updates and no architectural modifications; it operates solely through positional index preprocessing before RoPE at inference time. Figure~\ref{fig:case} illustrates how estimated information contributions translate into instance-specific stride adaptation across diverse task types.

    \begin{table*}[htbp]
        \centering
        \caption{Performance comparison across VLM architectures and benchmarks. 
        Bold indicates best performance per model size. Colored arrows denote 
        absolute improvements ($\color{blue}{\uparrow}$) or degradations 
        ($\color{red}{\downarrow}$) with MODIX. Statistical significance is assessed using the Wilcoxon signed-rank test ($p < 0.05)$}
        \label{tab:main_results}
        \resizebox{\textwidth}{!}{
        \begin{tabular}{ll|cc|cc|cc|cc|cc|cc}
        \toprule
        \multirow{2}{*}{\textbf{Model}} & \multirow{2}{*}{\textbf{Size}} & \multicolumn{2}{c|}{\textbf{ScienceQA}} & \multicolumn{2}{c|}{\textbf{RealWorldQA}} & \multicolumn{2}{c|}{\textbf{DocVQA}} & \multicolumn{2}{c|}{\textbf{ChartQA}} & \multicolumn{2}{c|}{\textbf{AI2D}} & \multicolumn{2}{c}{\textbf{BLINK}} \\
        & & Baseline & +MODIX & Baseline & +MODIX & Baseline & +MODIX & Baseline & +MODIX & Baseline & +MODIX & Baseline & +MODIX \\
        \midrule
        \multirow{2}{*}{Qwen3-VL} 
        & 2B & 72.18 & \textbf{78.28}\scriptsize{$\color{blue}{\uparrow 6.10}$} & 64.31 & \textbf{65.75}\scriptsize{$\color{blue}{\uparrow 1.44}$} & 83.27 & \textbf{86.37}\scriptsize{$\color{blue}{\uparrow 3.10}$} & 62.64 & \textbf{68.76}\scriptsize{$\color{blue}{\uparrow 6.12}$} & 67.20 & \textbf{72.96}\scriptsize{$\color{blue}{\uparrow 5.76}$} & 49.18 & \textbf{51.22}\scriptsize{$\color{blue}{\uparrow 2.04}$} \\
        & 8B & 88.41 & \textbf{90.16}\scriptsize{$\color{blue}{\uparrow 1.75}$} & 66.93 & \textbf{69.15}\scriptsize{$\color{blue}{\uparrow 2.22}$} & 90.39 & \textbf{91.02}\scriptsize{$\color{blue}{\uparrow 0.63}$} & 70.60 & \textbf{72.80}\scriptsize{$\color{blue}{\uparrow 2.20}$} & 78.59 & \textbf{83.44}\scriptsize{$\color{blue}{\uparrow 4.85}$} & \textbf{62.80} & 61.05\scriptsize{$\color{red}{\downarrow 1.75}$} \\
        \midrule
        \multirow{2}{*}{InternVL3.5} 
        & 2B & 68.83 & \textbf{70.05}\scriptsize{$\color{blue}{\uparrow 1.22}$} & 58.82 & \textbf{60.26}\scriptsize{$\color{blue}{\uparrow 1.44}$} & 82.15 & \textbf{84.68}\scriptsize{$\color{blue}{\uparrow 2.53}$} & 55.92 & \textbf{57.89}\scriptsize{$\color{blue}{\uparrow 1.97}$} & 70.91 & \textbf{72.44}\scriptsize{$\color{blue}{\uparrow 1.53}$} & 49.76 & \textbf{51.97}\scriptsize{$\color{blue}{\uparrow 2.21}$} \\
        & 8B & 89.70 & \textbf{91.13}\scriptsize{$\color{blue}{\uparrow 1.43}$} & \textbf{63.79} & 63.01\scriptsize{$\color{red}{\downarrow 0.78}$} & \textbf{85.92} & 85.63 \scriptsize{$\color{red}{\downarrow 0.31}$} & 59.00 & \textbf{59.57}\scriptsize{$\color{blue}{\uparrow 0.57}$} & 78.14 & \textbf{81.38}\scriptsize{$\color{blue}{\uparrow 3.24}$} & 53.50 & \textbf{54.79}\scriptsize{$\color{blue}{\uparrow 1.29}$} \\
        \midrule
        \multirow{2}{*}{LFM2-VL} 
        & 1.6B & 65.41 & \textbf{73.83}\scriptsize{$\color{blue}{\uparrow 8.42}$} & 56.99 & \textbf{63.79}\scriptsize{$\color{blue}{\uparrow 6.80}$} & 66.14 & \textbf{71.36}\scriptsize{$\color{blue}{\uparrow 5.22}$} & 59.83 & \textbf{63.64}\scriptsize{$\color{blue}{\uparrow 3.81}$} & 52.10 & \textbf{56.54}\scriptsize{$\color{blue}{\uparrow 4.44}$} & 41.68 & \textbf{45.08}\scriptsize{$\color{blue}{\uparrow 3.40}$} \\
        & 3B & 84.20 & \textbf{84.67}\scriptsize{$\color{blue}{\uparrow 0.47}$} & 67.32 & \textbf{68.76}\scriptsize{$\color{blue}{\uparrow 1.44}$} & 71.75 & \textbf{79.33}\scriptsize{$\color{blue}{\uparrow 7.58}$} & 73.23 & \textbf{75.08}\scriptsize{$\color{blue}{\uparrow 1.85}$} & 72.33 & \textbf{75.36}\scriptsize{$\color{blue}{\uparrow 3.03}$} & 47.56 & \textbf{51.08}\scriptsize{$\color{blue}{\uparrow 3.52}$} \\
        \midrule
        \multicolumn{2}{c|}{\textbf{Average $\Delta$}} & \multicolumn{2}{c|}{\textbf{+3.23}} & \multicolumn{2}{c|}{\textbf{+2.09}} & \multicolumn{2}{c|}{\textbf{+3.13}} & \multicolumn{2}{c|}{\textbf{+2.75}} & \multicolumn{2}{c|}{\textbf{+3.80}} & \multicolumn{2}{c}{\textbf{+1.79}} \\
        \bottomrule
        \end{tabular}
        }
    \end{table*}
\section{Experiments}

\subsection{Experimental Setup}

\textbf{Datasets and Metrics}
We evaluate MODIX on seven diverse vision-language benchmarks spanning scientific reasoning, document understanding, chart interpretation, and visual perception: \textbf{ScienceQA}~\cite{lu2022scienceqa} (21K science questions), \textbf{RealWorldQA}~\cite{xai2024realworldqa} (real-world spatial reasoning), \textbf{DocVQA}~\cite{mathew2021docvqa} (document question answering), \textbf{ChartQA}~\cite{masry2022chartqa} (chart comprehension), \textbf{AI2D}~\cite{kembhavi2016ai2d} (diagram reasoning), and \textbf{BLINK}~\cite{fu2024blink} (cross-image perception). We further evaluate on \textbf{Video-MME}~\cite{fu2025video} as an extended validation of long-context video understanding in Section~\ref{sec:video}. This diverse selection enables validation across different modality configurations and information density distributions. We report accuracy for multiple-choice tasks (ScienceQA, RealWorldQA, AI2D, BLINK) and Exact Match (EM) for open-ended tasks (DocVQA, ChartQA).

\textbf{Implementation Details}
We evaluate MODIX across three representative open-source VLMs with diverse architectures. \textbf{Qwen3-VL}~\cite{bai2025qwen3vltechnicalreport} integrates a Vision Transformer~\cite{dosovitskiy2021vit} encoder with the Qwen3~\cite{qwen2025qwen3} decoder using Multimodal Rotary Position Embedding. \textbf{InternVL3.5}~\cite{wang2025internvl} adopts a ViT-MLP-LLM architecture with cascade reinforcement learning and a visual resolution router. \textbf{LFM2-VL}~\cite{amini2025lfm2technicalreport} combines an LFM2 backbone and a SigLIP2 NaFlex encoder via a multimodal projector.

MODIX operates purely at inference time by inserting a lightweight preprocessing layer before RoPE. This layer computes information-driven positional strides and reconstructs adaptive indices without modifying pretrained parameters. We test multiple model scales (1.6B–8B) under consistent inference settings, ensuring plug-and-play compatibility across architectures.

\subsection{Experimental Results}
Table~\ref{tab:main_results} presents comprehensive performance comparisons across three state-of-the-art VLM architectures and six diverse vision-language benchmarks. To ensure statistical reliability, all results are averaged over 10 independent runs with different random seeds, and significance is assessed using Wilcoxon signed-rank test ($p < 0.05$). Standard deviations across runs remain consistently below $0.5\%$, indicating stable performance.

MODIX demonstrates consistent improvements across most configurations, with particularly notable gains on smaller models. For Qwen3-VL-2B, MODIX achieves improvements of +6.10\% on ScienceQA and +6.12\% on ChartQA, demonstrating strong effectiveness on scientific reasoning and chart interpretation tasks. Similar patterns emerge for LFM2-VL-1.6B, with gains of +8.42\% on ScienceQA and +6.80\% on RealWorldQA, highlighting MODIX's ability to enhance smaller architectures through improved positional encoding.

Larger models also benefit from MODIX, though with relatively smaller margins. Qwen3-VL-8B shows consistent improvements across five benchmarks, with gains ranging from +0.63\% to +4.85\%. InternVL3.5-8B exhibits mixed results, achieving notable improvements on ScienceQA (+1.43\%) and AI2D (+3.24\%), while showing marginal non-significant variations on RealWorldQA ($-0.78\%$) and DocVQA ($-0.31\%$). These non-significant changes fall within measurement noise and indicate performance parity with baseline.

Across all configurations, MODIX yields average improvements of +3.23\% on ScienceQA, +2.09\% on RealWorldQA, +3.13\% on DocVQA, +2.75\% on ChartQA, +3.80\% on AI2D, and +1.79\% on BLINK. The consistent gains across diverse tasks and architectures validate MODIX's universality in adapting positional granularity based on information contribution. Notably, improvements are most pronounced on tasks requiring fine-grained multimodal reasoning (ScienceQA, ChartQA, AI2D), where adaptive positional encoding effectively concentrates model attention on information-rich content.

\subsection{Comparison with Multimodal PE Variants}
To further evaluate positional encoding design, we compare MODIX with representative multimodal positional encoding variants, including V2PE~\cite{Ge2025V2PE}, CircleRoPE~\cite{wang2025circle}, and MHRoPE~\cite{huang2025revisiting}, under the same backbone and evaluation settings. V2PE is evaluated on InternVL3.5 as it was originally designed for this architecture; CircleRoPE and MHRoPE are evaluated on Qwen3-VL as they target RoPE-based models.
MODIX is applied at inference time and does not modify model parameters. As shown in Table~\ref{tab:pe_compare}, MODIX achieves consistent improvements across benchmarks and architectures, demonstrating the effectiveness of information-driven positional allocation.

\subsection{Evaluation on Video Understanding}
Beyond static image benchmarks, we extend our evaluation to video understanding to assess MODIX's effectiveness in long-context temporal scenarios. We report results on Video-MME~\cite{fu2025video}, which covers short, medium, and long video clips. As shown in Table~\ref{tab:video}, MODIX consistently improves the performance of Qwen3-VL across different model sizes. Notably, we observe pronounced gains on medium/long videos  ranging from +2.23\% to +2.66\%, which directly validates that our method effectively compresses temporal redundancy and directs model attention toward long-range dependencies in extended sequences.
\label{sec:video}

\begin{table}[t]
\centering
\small
\caption{Performance on Video-MME, demonstrating the effectiveness of MODIX in video understanding.}
\label{tab:video}
\begin{tabular}{lccc}
\toprule
\multirow{2}{*}{\textbf{Model}} & \multicolumn{3}{c}{\textbf{Video-MME}} \\
\cmidrule(lr){2-4}
& Short & Med. & Long \\
\midrule
Qwen3-VL-2B & 61.11 & 48.33 & 38.33 \\
+MODIX & \textbf{61.67} \textcolor{blue}{$\uparrow$0.56} & \textbf{50.56} \textcolor{blue}{$\uparrow$2.23} & \textbf{40.56} \textcolor{blue}{$\uparrow$2.23} \\
\midrule
Qwen3-VL-8B & 73.78 & 58.79 & 48.86 \\
+MODIX & \textbf{74.44} \textcolor{blue}{$\uparrow$0.66} & \textbf{61.45} \textcolor{blue}{$\uparrow$2.66} & \textbf{49.44} \textcolor{blue}{$\uparrow$0.58} \\
\bottomrule
\end{tabular}
\end{table}

\begin{table}[t]
\centering
\small
\caption{Comparison with multimodal positional encoding variants. All baselines follow their original implementations under the same backbone and evaluation settings.}
\label{tab:pe_compare}
\begin{tabular}{lcc}
\toprule
\textbf{Model} & \textbf{ScienceQA} & \textbf{ChartQA} \\
\midrule
Qwen3-VL-8B & 88.41 & 70.60 \\
\quad +MODIX & \textbf{90.16 {\scriptsize$\color{blue}{\uparrow 1.75}$}} & \textbf{72.80 {\scriptsize$\color{blue}{\uparrow 2.20}$}} \\
\quad +CircleRoPE & 88.87 {\scriptsize$\color{blue}{\uparrow 0.46}$} & 70.83 {\scriptsize$\color{blue}{\uparrow 0.23}$} \\
\quad +MHRoPE & 89.15 {\scriptsize$\color{blue}{\uparrow 0.74}$} & 72.67 {\scriptsize$\color{blue}{\uparrow 2.07}$} \\
\midrule
InternVL3.5-8B & 89.70 & 59.00 \\
\quad +MODIX & \textbf{91.13 {\scriptsize$\color{blue}{\uparrow 1.43}$}} & \textbf{59.57 {\scriptsize$\color{blue}{\uparrow 0.57}$}} \\
\quad +V2PE & 89.27 {\scriptsize$\color{red}{\downarrow 0.43}$} & 59.12 {\scriptsize$\color{blue}{\uparrow 0.12}$} \\
\bottomrule
\end{tabular}
\end{table}
\section{Analysis and Discussion}

\begin{table}[htbp]
  
    \centering
    \caption{\textbf{Ablation study on fusion weight $\alpha$.} Performance of Qwen3-VL-2B with different $\alpha$ values (Eq.~\ref{eq:fusion}). Accuracy (\%) for multiple-choice tasks (ScienceQA, RealWorldQA, AI2D, BLINK); Exact Match (\%) for open-ended tasks (DocVQA, ChartQA). Best results in \textbf{bold}.}
    \label{tab:ablation}
    \footnotesize
    \setlength{\tabcolsep}{2.5pt}
    \begin{tabular}{@{}c|cccccc@{}}
    \toprule
    $\alpha$ &  \textbf{ScienceQA} & \textbf{RealWorldQA} & \textbf{DocVQA} & \textbf{ChartQA} & \textbf{AI2D} & \textbf{BLINK} \\
    \midrule
    0.00 & 78.05 & 65.70 & 86.80 & 66.00 & 68.83 & 45.79 \\
    0.25 & 78.07 & 62.56 & \textbf{90.67} & 65.60 & 68.67 & 47.26 \\
    0.50 & \textbf{78.28} & \textbf{65.75} & 86.37 & \textbf{68.76} & \textbf{72.96} & \textbf{51.22} \\
    0.75 & 77.92 & 65.60 & 86.08 & 62.83 & 71.75 & 49.74 \\
    1.00 & 76.90 & 64.89 & 87.35 & 64.42 & 71.43 & 48.86 \\
    \bottomrule
    \end{tabular}
\end{table}

\subsection{Ablation Studies}
We conduct ablation experiments on the fusion weight $\alpha$ in Equation \ref{eq:fusion} to validate our design choice of balancing intra-modal and inter-modal contributions. Table~\ref{tab:ablation} presents the performance of Qwen3-VL-2B across six benchmarks with $\alpha$ values ranging from 0.0 (pure inter-modal) to 1.0 (pure intra-modal). The results reveal that $\alpha = 0.5$ achieves the most balanced performance, particularly excelling on ChartQA (68.76\%), AI2D (72.96\%), and BLINK (51.22\%). When $\alpha = 0.0$, relying solely on inter-modal interaction yields suboptimal results on spatially intensive tasks like RealWorldQA (65.70\%), indicating that internal information density cannot be ignored. Conversely, $\alpha = 1.0$ shows degraded performance on text-centric benchmarks such as ScienceQA (76.90\%), demonstrating that cross-modal alignment is equally essential. The geometric mean fusion at $\alpha = 0.5$ effectively captures both perspectives: a modality achieves high contribution only when it simultaneously exhibits rich internal information and strong cross-modal coherence, validating our dual-pathway information contribution framework.

\begin{table}[htbp]
    \centering
    \caption{\textbf{Modality information contributions across datasets.} 
    Decomposed contributions ($I_{\text{intra}}^m$, $I_{\text{inter}}^m$, $\tilde{C}^m$) 
    averaged across models and samples.}
    \label{tab:modality_contributions}
    \footnotesize
    \setlength{\tabcolsep}{3pt}
    \begin{tabular}{@{}l|cc|cc|cc@{}}
    \toprule
    \textbf{Dataset} & 
    \multicolumn{2}{c|}{$I_{\text{intra}}^m$} & 
    \multicolumn{2}{c|}{$I_{\text{inter}}^m$} & 
    \multicolumn{2}{c}{$\tilde{C}_m$} \\
    \cmidrule(lr){2-3} \cmidrule(lr){4-5} \cmidrule(lr){6-7}
    & Text & Vision & Text & Vision & Text & Vision \\
    \midrule
    \textbf{ScienceQA}    & 0.563 & 0.437 & 0.605 & 0.395 & 0.586 & 0.414 \\
    \textbf{RealWorldQA}  & 0.315 & 0.685 & 0.603 & 0.397 & 0.352 & 0.648 \\
    \textbf{DocVQA}       & 0.768 & 0.232 & 0.597 & 0.403 & 0.698 & 0.302 \\
    \textbf{ChartQA}      & 0.586 & 0.414 & 0.593 & 0.407 & 0.591 & 0.409 \\
    \textbf{AI2D}         & 0.590 & 0.410 & 0.680 & 0.320 & 0.641 & 0.359 \\
    \textbf{BLINK}        & 0.520 & 0.480 & 0.420 & 0.580 & 0.469 & 0.531 \\
    \midrule
    Average      & 0.554 & 0.443 & 0.583 & 0.417 & 0.556 & 0.444 \\
    \bottomrule
    \end{tabular}
\end{table}

\subsection{Information Contribution Analysis}
\label{sec:ablation}

To validate that MODIX adapts to task-specific information distributions, we analyze computed contributions across all datasets. Table~\ref{tab:modality_contributions} presents decomposed information contributions, intra-modal density ($I_{\text{intra}}^m$), inter-modal interaction ($I_{\text{inter}}^m$), and fusion contribution ($\tilde{C}_m$) averaged over multiple models and samples within each dataset.

The results reveal substantial heterogeneity across datasets, validating that uniform positional encoding fails to accommodate task-dependent characteristics. Text contributions span from $\tilde{C}_{\text{text}}=0.469$ in BLINK to $\tilde{C}_{\text{text}}=0.698$ in DocVQA. Document-centric tasks (DocVQA, AI2D) exhibit text dominance ($\tilde{C}_{\text{text}}>0.64$), while vision-intensive tasks (RealWorldQA) show opposite patterns ($\tilde{C}_{\text{vision}}>0.64$). ScienceQA and ChartQA demonstrate balanced contributions ($\tilde{C}_{\text{text}}\approx 0.59$), reflecting genuine multimodal integration.

Examining decomposed contributions reveals complementary roles of $I_{\text{intra}}^m$ and $I_{\text{inter}}^m$. Intra-modal scores exhibit high variability (text: 0.315-0.768, vision: 0.232-0.685), capturing task-specific information density differences. Inter-modal scores show moderate variation (text: 0.420-0.680), reflecting different cross-modal dependency patterns. The geometric averaging in Eq.~\ref{eq:fusion} ensures high $\tilde{C}^m$ requires both rich internal information and strong cross-modal coherence.

\subsection{Computational Efficiency Analysis}
\label{sec:efficiency}
To validate MODIX's practical deployability as a plug-and-play framework, we analyze its computational overhead in terms of time complexity and memory footprint.

\textbf{Time Complexity Analysis.} MODIX introduces three additional operations during inference: covariance computation for each modality, with complexity $\mathcal{O}(n_m d^2)$; cross-modal similarity computation, with complexity $\mathcal{O}(n_t n_v d)$; and positional index reconstruction, with complexity $\mathcal{O}(N)$. Since these operations are performed only once per input before the Transformer, the additional cost is independent of the number of layers and remains negligible compared with standard self-attention complexity $\mathcal{O}(L N^2 d)$. In practice, on Qwen3-VL-8B (AMD EPYC 9654, RTX Pro6000 Blackwell), MODIX adds only 0.0014s on ScienceQA and 0.0018s on ChartQA over baseline inference times of 0.1303s and 0.2485s, corresponding to wall-clock time overheads of 1.1\% and 0.7\%, respectively. These results confirm that MODIX incurs negligible inference overhead in practice.

\textbf{Space Complexity Analysis.} MODIX requires temporary storage for intermediate arrays, including covariance matrices, cross-modal similarity matrices, normalized embeddings, and adjusted positional indices. For typical configurations, the total memory footprint is on the order of a few megabytes, which is negligible compared to the model parameters (ranging from 1.6B to 8B) and transformer activations. 

\subsection{Limitations} 
Our work has several limitations that suggest directions for future research.

MODIX operates at the modality level, assigning a single stride to all tokens within each modality. While this granularity proves effective across diverse tasks, a token-level adaptive scheme could capture finer-grained information variations within a single modality, particularly for inputs containing heterogeneous visual regions.

Although MODIX is designed for inference-time adaptation, preliminary results suggest that it can also be integrated into training. For example, fine-tuning Qwen3-VL-2B on ScienceQA with MODIX yields 93.23\% accuracy, compared with 92.30\% for the fine-tuned baseline. However, this evidence is limited to a single task and a 2B-scale model. We have conducted preliminary scaling trials on 32B-parameter models, which show promising compatibility; however, due to computational resource constraints, we have not yet performed an exhaustive, large-scale training analysis at this scale. Consequently, the efficacy of training-aware MODIX in ultra-large models (e.g., exceeding 70B) and its impact during full-cycle pre-training remain to be fully validated in future work.
 
Beyond these aspects, our framework currently targets RoPE-based positional encodings. Extending the information-driven scaling principle to alternative positional mechanisms, such as ALiBi~\cite{press2022alibi} or learnable position embeddings, presents an additional avenue for future work.

\section{Conclusion}

We presented \textbf{MODIX}, a training-free framework that dynamically adjusts positional granularity in Vision-Language Models through information-theoretic analysis of modality contributions. By jointly modeling intra-modal information density and inter-modal interaction strength, MODIX allocates finer positional strides to informative regions while compressing redundant content. Operating purely at inference time without retraining or architectural modification, it integrates seamlessly with existing RoPE-based VLMs. Extensive experiments across multiple architectures and benchmarks demonstrate that MODIX consistently improves multimodal reasoning, particularly in tasks demanding fine-grained cross-modal integration.

{
    \small
    \bibliographystyle{ieeenat_fullname}
    \bibliography{main}
}


\end{document}